\pdfoutput=1
\documentclass[11pt]{article}

\usepackage[preprint]{acl}

\usepackage{times}
\usepackage{latexsym}

\usepackage[T1]{fontenc}
\usepackage[utf8]{inputenc}

\usepackage{microtype}

\usepackage{inconsolata}

\usepackage{graphicx}

\usepackage{booktabs}

\title{Automated Evaluation of Meter and Rhyme in Russian Generative and Human-Authored Poetry}

\author{Ilya Koziev \\
  \small{
    \textbf{Correspondence:} \href{inkoziev@gmail.com}{inkoziev@gmail.com}
  }
}

\begin{document}
\maketitle
\begin{abstract}

Generative poetry systems require effective tools for data engineering and automatic evaluation, particularly to assess how well a poem adheres to versification rules, such as the correct alternation of stressed and unstressed syllables and the presence of rhymes.

In this work, we introduce the \texttt{Russian Poetry Scansion Tool} library designed for stress mark placement in Russian-language syllabo-tonic poetry, rhyme detection, and identification of defects of poeticness. Additionally, we release \texttt{RIFMA} -- a dataset of poem fragments spanning various genres and forms, annotated with stress marks and rhyme schemes. This dataset can be used to evaluate the capability of modern LLMs to accurately place stress marks in poetic texts.

The published resources provide valuable tools for researchers and practitioners in the field of creative generative AI, facilitating advancements in the development and evaluation of generative poetry systems.

\end{abstract}

\section{Introduction}

Generative poetry, like other creative tasks, presents a compelling domain for AI research~\cite{colton2012computational}, capturing not only academic interest but also the imagination of the general public~\cite{porter2024ai}. A critical challenge in creative computing, however, lies in the evaluation of generated outputs~\cite{gomez2023confederacy}. While human evaluation is often necessary, it is costly, time-consuming, and difficult to scale, limiting the pace and scope of research. Fortunately, certain formal aspects of poetry — such as meter and rhyme — lend themselves well to computational analysis. These properties have even been leveraged to develop benchmarks for evaluating LLMs~\cite{walsh2024sonnet}. In this paper, we address this opportunity by introducing a novel approach to the computational assessment of poetic meter and rhyme, with a focus on Russian-language syllabo-tonic poetry. Our method has been rigorously tested and proven effective in practice.

Our key contributions are as follows:

\begin{itemize}
    \item A new test dataset (\ref{sec:test_dataset}) comprising approximately 5,100 Russian human-authored poetry stanzas, each annotated with stress marks and rhyme scheme information. This dataset serves as a valuable resource for both research and benchmarking.
    \item The \texttt{Russian Poetry Scansion Tool} (RPST), an open-source library for the technical analysis of Russian poetry, made publicly available in our repository (\ref{sec:repo}).
\end{itemize}

By addressing the challenges of formal poetic analysis and providing practical tools and datasets, our work advances the field of computational creativity and opens new avenues for research in generative poetry.

\section{Related Work}

Scansion and stress assignment are critical components of poetry generation systems, ensuring that generated texts adhere to the metrical and rhythmic patterns of the target language. For example, \citet{ram2021say} employ phonetic transcription libraries such as eSpeak\footnote{\url{http://espeak.sourceforge.net/}} and Festival\footnote{\url{https://www.cstr.ed.ac.uk/projects/festival/}} to generate pop lyrics. Many English-language systems rely on the CMU Pronouncing Dictionary\footnote{\url{http://www.speech.cs.cmu.edu/cgi-bin/cmudict}} for stress placement, as demonstrated by \citet{agirrezabal2023erato}. However, the effectiveness of this tool is limited due to its incomplete coverage and inability to account for the prosodic variability of English. To address this limitation, \citet{greene-etal-2010-automatic} propose an unsupervised approach to stress placement using finite-state automata (FSA). \citet{ghazvininejad2016generating} provide a detailed discussion of stress placement, emphasizing the importance of secondary stress (included in the CMU Pronouncing Dictionary) and the challenges of adjusting stress patterns in poetic texts. For Spanish-language poetry, \citet{Marco2021AutomatedMA} present algorithmic approaches to stress assignment. A comprehensive review of tools for annotating poetic texts in Indo-European languages, as well as Finnish and Basque, is provided by \citet{Sisto2024UnderstandingPU}, including links to relevant code repositories.

% тут про оценку техничности в сгенерированных стихах:
Automatic evaluation of technical aspects of poetry is employed in various projects.
For example, \citet{possi2023carmen} propose metrics for assessing rhyme and meter automatically.
\citet{zhao2022automatic} introduce a ``tone-checker'' module with rules tailored to classical Chinese poetry.
\citet{chudoba2024gpt} use a set of automatic metrics to evaluate the form of generated poems, including compliance with meter, rhyme, and syllable count.
Similarly, \citet{hu2024poetrydiffusion} incorporate a metrical controller to constrain their diffusion-based poetry model.
\citet{nguyen2021sp} apply automatic quantitative evaluation to assess rhyme and tone rule conformance in Vietnamese poetry generation.
\citet{agirrezabal2023erato} describe a suite of automated tests that evaluate poeticness (e.g., rhyme richness), novelty (using \texttt{ROUGE} as a similarity metric), and topicality (using BERT-based embeddings).
Poeticness was one of the criteria in human assessment of generated poetry described in \citet{khanmohammadi2023prose2poem,van2020automatic,chen2019sentiment,yan2016poet,wang2016chinese}.

% далее про детектирование рифмовки
\textit{Rhyme} is another essential feature of poetic texts and song lyrics. For generative poetry systems based on large language models, a high-quality training dataset is crucial, as the model learns rhyme patterns during fine-tuning. The quality and completeness of such datasets are critical for generating high-quality poetry. Therefore, at the data engineering stage, tools are needed to detect poor or missing rhymes, enabling the removal of such samples from the dataset. These tools are also valuable for the automatic evaluation of generated poems, as the absence of rhymes significantly reduces the quality of the output.

Given these requirements, various approaches to rhyme detection have been explored in the field of generative poetry. \citet{ghazvininejad2016generating} provide a detailed description of the process of selecting exact and slant rhymes for sonnets. \citet{hirjee2009automatic} describe an algorithmic approach to detecting partial and internal rhymes in rap lyrics. A neural network-based approach for English, German, and French poetry is presented in \citet{haider2018supervised}. \citet{reddy2011unsupervised} discuss the compilation of a rhyme dictionary using n-gram statistics. Additionally, \citet{haider-2021-metrical} explore corpus-driven neural models that model the prosodic features of syllables and evaluate against rhythmically diverse data, considering both syllable-level and line-level features.

\section{Problem Definition}
\label{sec:problem-definition}

This paper addresses the development of a tool for automatically evaluating two key features of Russian syllabo-tonic poetry: (1) poetic meter and (2) rhyme quality. 

The tool must meet the following requirements:
\begin{itemize}
    \item \textbf{Accuracy}: The evaluation quality should be high enough to eliminate the need for expert intervention during the assessment process.
    \item \textbf{Scalability}: The tool should process millions of poems efficiently on standard hardware available to NLP researchers.
    \item \textbf{Extensibility}: The implementation should support easy integration of new lexicons and provide interpretable outputs to facilitate analysis and debugging.
\end{itemize}

\section{Russian Poetry Scansion Tool}

\texttt{RPST} is a python library designed to analyze and mark up texts of syllabo-tonic poems and songs in Russian. This library is available at the repository (\ref{sec:repo}). The tasks it performs include:

\begin{itemize}

    \item Stress placement in the text of a poem and song with adjustment to the poetic meter. Detection of poetic meter.

    \item Detection of prosodic defects and calculation of \textit{technicality}\footnote{We use the term \textit{technicality} rather than \textit{poeticness}~\cite{manurung2004evolutionary} to highlight the strictly formal nature of this assessment. The term \textit{poeticness} often carries broader, more subjective connotations.} - scores in the range from 0 (complete non-compliance with poetic constraints) to 1 (perfect compliance with one of the poetic meters).

    \item Detection of rhymes, including fuzzy ones (slant rhymes).
    
\end{itemize}

There are two primary use cases for \texttt{RPST}. First, during data engineering, \texttt{RPST} can filter out defective poems with poorly observed meter and rhyme from the training dataset. This scenario is detailed in Section~\ref{sec:amateur-poetry-assessment}. Second, for evaluating generated poems, \texttt{RPST} provides a quantitative measure of technicality, enabling the ranking of outputs before presenting them to the user. This scenario is discussed in Sections \ref{sec:human-assessment-neuropoetry} and \ref{sec:sbs}.

Key features of the algorithm underlying \texttt{RPST} include:

\begin{itemize}

\item Function words, such as prepositions, conjunctions, particles, are usually unstressed. This rule also applies to some auxiliary verbs.

\item In some cases, in a preposition + noun collocation, the stress moves to the preposition, leaving the noun unstressed. There are also a number of 3-word phrases that violate the usual stress rules of the Russian language.

\item Stresses are adjusted to the general poetic meter. This is expressed in the fact that a noun or a polysyllabic preposition will remain unstressed, although this is usually a sign of a defect. In the Russian language, there are many words with variable stress, when either the literary norm allows for variations, or one of the variations has a stylistic coloring. In these cases, the variation that minimizes the discrepancy with the poetic meter is chosen.

\item Out-of-vocabulary (OOV) words are processed by a set of rules that take into account the most popular word-formation prefixes, as well as by a small neural network model that predicts the stress position. You can also enable the option of adjusting the stress in OOV words simply to the poetic meter.

\item The total number of stress placement variants in each line is usually too large to check them by simple enumeration, so \texttt{RPST} uses beam search and a number of heuristics to limit the enumeration space. In most cases, this ensures fast and accurate text analysis.

\end{itemize}

\section{Human Evaluation}
\label{sec:human-eval}

\subsection{Human Assessment of Generative Poetry}
\label{sec:human-assessment-neuropoetry}

We used the results of a crowdsourced assessment of rhyme and meter in Russian-language poems generated by several large language models (LLMs). In total, 875 generated texts were evaluated. Most texts were rated by five non-expert annotators, while a small portion received four or three ratings. For consistency, 859 texts with a complete overlap of five ratings were included in the subsequent analysis.

Each text was assigned a score on a three-point scale: 0 (no rhyme, meter not observed), 1 (partial rhyme or meter), and 2 (rhyme present, meter observed). The distribution of assigned scores was as follows:

\begin{itemize}
    \item Score 0: 952 ratings
    \item Score 1: 1571 ratings
    \item Score 2: 1772 ratings
\end{itemize}

For correlation analysis, the mean score of each text was computed across all raters. Each text was then analyzed using the RPST library, resulting in two corresponding lists of real-valued outputs. The correlation between these lists was computed using the \texttt{scipy} library.

The correlation is statistically significant at $\alpha = 0.05$ (Pearson's $r = 0.79$, $p = 3.2 \times 10^{-181}$).

\subsection{Side-by-side Evaluation of Poetry}
\label{sec:sbs}

We conducted two sessions of side-by-side evaluation, comparing poems generated by different language models and those authored by humans. In each session, experts analyzed pairs of poems (generated or written from the same prompt) and selected the best in each pair based on several criteria: adherence to versification norms, grammatical correctness, coherence, relevance to the prompt, and poeticity. The two sessions differed significantly in composition: Session~1 included a larger number of models, some of which were a priori weaker, while Session~2 focused on two generative models of similar quality and human-authored poems.

Table~\ref{tab:sbs-technicality} presents the average technicality scores for these poems, calculated using \texttt{RPST}. The results show that the average technicality of the selected poems in each pair is higher than that of the losing poems. However, this does not imply that technicality alone correlates with the annotators' choices. To investigate this, we calculated the inter-rater agreement between the annotators' selections and a hypothetical selection based solely on technicality (i.e., choosing the poem with the higher technicality score in each pair). The results, shown in Table~\ref{tab:inter-rater-agreement}, indicate moderate agreement for Session~1, where the range of poem quality was wider, and weak agreement for Session~2, where the compared texts were of similar quality. This suggests that other factors beyond technicality influence the annotators' decisions, especially when evaluating poems of comparable quality.

\begin{table*}[h!]
\centering
\begin{tabular}{lccc}
\toprule

Session & Num. of pairs & Winning poem & Losing poem \\
\hline
1 & 3080 & 0.660 $\pm$ 0.008  &  0.392 $\pm$ 0.014 \\
2 & 795 & 0.580 $\pm$ 0.02 & 0.511 $\pm$ 0.021 \\
\bottomrule
\end{tabular}
\caption{Average technicality scores for winning and losing poems in side-by-side evaluation pairs. Error margins are calculated at a 95\% confidence level.}
\label{tab:sbs-technicality}
\end{table*}

\begin{table}[h!]
\centering
\begin{tabular}{lc}
\toprule

Session & Cohen's kappa \\
\hline
1 & 0.38 \\
2 & 0.18 \\
\bottomrule
\end{tabular}
\caption{Inter-rater agreement of human assessors and technicality-based choice for side-by-side sessions.}
\label{tab:inter-rater-agreement}
\end{table}

\section{Technical Assessment of Internet-Sourced Amateur Poetry}
\label{sec:amateur-poetry-assessment}

We have collected a corpus of approximately 18 million poems authored by amateurs. The source was websites where authors publish their works themselves, usually without moderation. The different levels of authors and the lack of an editorial filter lead to a large number of technically poor-quality texts. Including such poems in the training data for LM leads to suboptimal results, so automatic evaluation and filtering out defective samples is a necessary part of data engineering. One of the tasks of \texttt{RPST} is precisely this evaluation. Below we present the results obtained for the collected corpus.

Table~\ref{tab:amateur-meter} presents the distribution of poetic meters among the analyzed poems. Notably, nearly half of the poems are composed in iambic meter.

\begin{table}[h!]
\centering
\begin{tabular}{lcc}
\toprule

Meter      & Num. of samples & Share, \% \\
\hline

iambic        & 8373643           & 47.1   \\
trochee      & 4200089           & 23.6   \\
anapest    & 2255317           & 12.7   \\
amphibrach & 1318943           & 7.4    \\
dactyl    & 832586            & 4.7    \\
other~\footnote{This includes \textit{dolnik} and other non-regular stress patterns}  & 783649            & 4.5    \\

\bottomrule
\end{tabular}
\caption{Distribution of poetic meters in amateur poems.}
\label{tab:amateur-meter}
\end{table}

\begin{figure}[h!]
\includegraphics[width=0.4\textwidth]{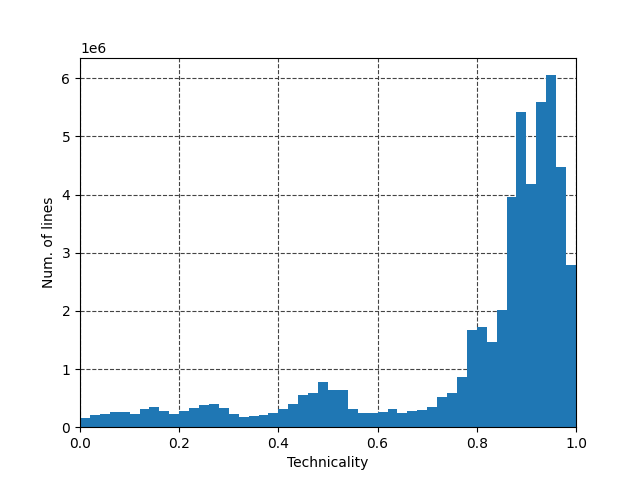}
\caption{Distribution of technicality scores per line for scraped poems.}
\label{fig:genitive_argument}
\end{figure}

Figure~\ref{fig:genitive_argument} shows the distribution of technicality scores for amateur poetry. The presence of multiple modes in the distribution, particularly around the score value of 0.45, warrants further investigation. We hypothesize that these modes may arise from specific genres or styles that do not fully align with the stress selection rules implemented in our analysis.

Table~\ref{tab:num-lines-above-threshold} shows the relationship between the cutoff threshold value and the number of lines with technicality scores above that threshold.

\begin{table}[h!]
\centering
\begin{tabular}{lcc}
\toprule

 Threshold & Num. of lines above & Share, \% \\
\hline
 0.7       & 41679096            & 78.5                    \\
 0.8       & 37687280            & 71.0                    \\
 0.9       & 23101285            & 43.5                    \\
\bottomrule
\end{tabular}
\caption{Number of lines with technicality scores above a given threshold.}
\label{tab:num-lines-above-threshold}
\end{table}

When selecting samples for the training dataset, a more informative criterion is the number of samples in which all lines have a technicality score above a given threshold. Table~\ref{tab:num-samples-above-threshold} summarizes these values for different thresholds. In practice, this means that fewer than one-third of the initial 18 million samples meet the criteria for inclusion in the training dataset.

\begin{table}[h!]
\centering
\begin{tabular}{lcc}
\toprule

 Threshold & Num. of poems above & Share, \% \\
\hline
 0.5       & 5055599             & 28.5                    \\
 0.6       & 3254027             & 18.3                    \\
 0.7       & 1623498             & 9.1                     \\
 0.8       & 694632              & 3.9                     \\
 0.9       & 208459              & 1.2                     \\
\bottomrule
\end{tabular}
\caption{Number of samples with technicality scores above a given threshold.}
\label{tab:num-samples-above-threshold}
\end{table}

\section{RIFMA: A Dataset for Russian Poetry with Accentuation Annotations}
\label{sec:test_dataset}

The \texttt{RIFMA} dataset consists of fragments of Russian-language poems (stanzas and poems) with stress marks and a rhyme scheme information. Fragments are taken from poems of different genres, forms and authored by different poets, to ensure the widest possible coverage. Statistics on sample length variations are presented in the Table~\ref{tab:num-lines-per-sample}.

\begin{table}[h!]
\centering
\begin{tabular}{lcc}
\toprule

Num. of lines & Num. of samples & Share, \% \\
\hline
4             & 2404            & 66.81    \\
3             & 830             & 23.07    \\
2             & 136             & 3.78     \\
1             & 85              & 2.36     \\
5             & 71              & 1.97     \\
8             & 30              & 0.83     \\
6             & 19              & 0.53     \\
12            & 17              & 0.47     \\
14            & 2               & 0.06     \\
16            & 1               & 0.03     \\
10            & 1               & 0.03     \\
9             & 1               & 0.03     \\
7             & 1               & 0.03     \\
\bottomrule
\end{tabular}
\caption{Distribution of poem lengths (in lines) in the \texttt{RIFMA} dataset.}
\label{tab:num-lines-per-sample}
\end{table}

\begin{table}[h!]
\centering
\begin{tabular}{lcc}
\toprule
Rhyme scheme   & Num. of samples & Share, \% \\
\hline

\texttt{ABAB}           & 659             & 18.32    \\
\texttt{-A-A}           & 604             & 16.79    \\
\texttt{ABBA}           & 596             & 16.56    \\
\texttt{AA-}            & 365             & 10.14    \\
\texttt{-{}-{}-}            & 222             & 6.17     \\
\texttt{A-A}            & 181             & 5.03     \\
\texttt{A-A-}           & 147             & 4.09     \\
\texttt{A--A}           & 122             & 3.39     \\
\texttt{-{}-{}-{}-}           & 114             & 3.17     \\
\texttt{-AA-}           & 107             & 2.97     \\

\bottomrule
\end{tabular}
\caption{Statistics on rhyme schemes in the \texttt{RIFMA} dataset. A dash (-) indicates unrhymed line. The table shows the top-10 most frequent variants.}
\label{tab:rifma-rhyming-schemes}
\end{table}

The dataset is available under MIT license in the repository (\ref{sec:repo}).

\section{Repository}
\label{sec:repo}

The \texttt{RPST} library is available as the repository \url{https://github.com/Koziev/RussianPoetryScansionTool}.

The \texttt{RIFMA} dataset is available at the repository \url{https://github.com/Koziev/Rifma}.

All of the above is published under the MIT license to promote collaboration and transparency in research. We encourage researchers and developers to use, modify, and build upon these resources to advance the field of computational poetry analysis.

\section{Conclusion and Future Work}

The approach to automatic poeticness assessment presented in this paper has demonstrated practical utility, particularly in improving the quality of training data and ranking outputs for a Russian-language poetry generation system. While the results are promising, there are several directions for future work to enhance the method's robustness and applicability.

First, the potential of using language models to address the challenges outlined in Section~\ref{sec:problem-definition} should be explored. This includes investigating their effectiveness in multilingual settings, which could broaden the approach's applicability beyond Russian-language poetry.

Second, practical utilization of \texttt{RPST} on web-collected data has highlighted the need for improved part-of-speech analysis. Specifically, the system must better handle grammatical, spelling, and punctuation errors, as well as the unique lexical and stylistic features characteristic of poetic texts. Addressing these issues would increase the method's reliability in real-world scenarios.

\section*{Limitations}

The \texttt{RPST} library is designed to handle the primary poetic meters of Russian syllabo-tonic verse, as well as some modern forms without a strict meter. To maintain practicality, we prioritized effective performance on the most common cases over rare or exceptional ones. As a result, some correctly written poems may produce incorrect results when analyzed by \texttt{RPST}.

The \texttt{RIFMA} dataset, in its current form, does not cover all poetic forms, genres, or expressive techniques used by poets. However, we plan to expand it incrementally to address these gaps over time.

\section*{Ethical Considerations}

\textbf{Responsible usage.} The development and use of the \texttt{RIFMA} dataset and the \texttt{RPST} library do not raise any significant ethical concerns. Both resources are designed to support research in computational poetry and are free from biases or harmful applications. However, as with any NLP tool, we encourage users to apply these resources responsibly and in ways that align with ethical research practices.

\textbf{AI-Assisted Writing.} This paper was proofread and improved using the DeepSeek assistant to correct grammatical, spelling, and stylistic errors, as well as to enhance readability. As a result, certain portions of the text may be flagged as AI-generated, AI-edited, or human-AI co-authored by detection tools. However, all ideas, research, and contributions remain entirely our own.

% Bibliography entries for the entire Anthology, followed by custom entries
%\bibliography{anthology,custom}
% Custom bibliography entries only
\bibliography{custom}

\appendix

%\section{Example Appendix}
%\label{sec:appendix}

%This is an appendix.

\end{document}